\documentclass{article}

\usepackage{iclr2023_workshop,times}

\usepackage{amsmath,amsfonts,bm}

\def\eqref#1{equation~\ref{#1}}

\def\1{\bm{1}}

\DeclareMathAlphabet{\mathsfit}{\encodingdefault}{\sfdefault}{m}{sl}
\SetMathAlphabet{\mathsfit}{bold}{\encodingdefault}{\sfdefault}{bx}{n}

\newcommand\diff{\mathrm{d}}

\usepackage{mathtools}
\usepackage{float}
\usepackage{amsmath, amsthm}
\usepackage{amssymb}
\usepackage{multirow}
\usepackage{pict2e}
\usepackage{caption}
\usepackage{subcaption}
\usepackage{stmaryrd}

\usepackage{natbib}
\setcitestyle{authoryear,sort}
\iclrfinalcopy

\newtheorem{theorem}{Theorem}[section]
 
\newtheorem{definition}{Definition}[section]
\newtheorem{lemma}{Lemma}[section]
\DeclarePairedDelimiter{\norm}{\lVert}{\rVert}
\usepackage{comment}
\usepackage{appendix}
\usepackage{xspace}
\usepackage{tikz}
\usepackage{makecell}
\usepackage{bm}
\usepackage{xcolor}

\usepackage[utf8]{inputenc} 
\usepackage[T1]{fontenc}    
\usepackage[colorlinks=true, allcolors=blue]{hyperref}     
\usepackage{url}            
\usepackage{booktabs}       
\usepackage{amsfonts}       
\usepackage{nicefrac}       
\usepackage{microtype}      

\newcommand{\ie}{i.e.\@\xspace}
\newcommand{\ts}{\textsuperscript}

\definecolor{julien}{rgb}{1, 0, 0}

\title{Stability of implicit neural networks for long-term forecasting in dynamical systems}

\author{Léon Migus\textsuperscript{1,2,3}, Julien Salomon\textsuperscript{2, 3}, Patrick Gallinari\textsuperscript{1,4} \\
\textsuperscript{1} Sorbonne Université, CNRS, ISIR, F-75005 Paris, France\\
\textsuperscript{2} INRIA Paris, ANGE Project-Team, 75589 Paris Cedex 12, France\\
\textsuperscript{3} Sorbonne Université, CNRS, Laboratoire Jacques-Louis Lions, 75005 Paris, France\\
\textsuperscript{4} Criteo AI Lab, Paris, France\\
\texttt{leon.migus@sorbonne-universite.fr, julien.salomon@inria.fr}\\\texttt{patrick.gallinari@sorbonne-universite.fr}
}

\begin{document}
\maketitle

\begin{abstract}
    Forecasting physical signals in long time range is among the most challenging tasks in Partial Differential Equations (PDEs) research. To circumvent limitations of traditional solvers, many different Deep Learning methods have been proposed. They are all based on auto-regressive methods and exhibit stability issues. Drawing inspiration from the stability property of implicit numerical schemes, we introduce a stable  auto-regressive implicit neural network. We develop a theory based on the stability definition of schemes to ensure the stability in forecasting of this network. It leads us to introduce hard constraints on its weights and propagate the dynamics in the latent space. Our experimental results validate our stability property, and show improved results at long-term forecasting for two transports PDEs.
\end{abstract}

\section{Introduction and motivation}
Numerical simulations are one of the main tools to study systems described by PDEs, which are essential for many applications including, e.g., fluid dynamics and climate science. 
However, solving these systems and even more using them to predict
long term phenomenon remains a complex challenge, mainly due to the accumulation of errors over time. To overcome the limitations of traditional solvers and to exploit the available data, many different deep learning methods have been proposed.

For the task of forecasting spatio-temporal dynamics, \citet{ayed2019learning} used a standard residual network with convolutions and \citet{sorteberg2019approximating, lino2020simulating, fotiadis2020comparing} used Long short-term memory (LSTM) and Convolutional neural network (CNN) for the wave equation. In \citet{wiewel2019latent, kim2019deep}, a good performances is obtained by predicting within the latent spaces of neural networks. More recently, \citet{brandstetter2022message} used graph neural networks with several tricks and showed great results for forecasting PDEs solutions behavior. Importantly, these methods all solve the PDE iteratively, meaning that they are auto-regressive, the output of the model is used as the input for the next time step. Another line of recent methods that have greatly improved the learning of PDE dynamics are Neural Operators \citep{li2020fourier}. These methods can be used as operators or as auto-regressive methods to forecast. However, when used as operators, they do not generalize well beyond the times seen during training.
Crucially, these auto-regressive methods tend to accumulate errors over time with no possible control, and respond quite poorly in case of change in the distribution of the data. This leads to stability problems, especially over long periods of time beyond the training horizon. 

In the numerical analysis community, the stability issue has been well studied and is usually dealt with implicit schemes. By definition, they imply to solve an equation to go from a time step to the next one but they are generally more stable than explicit schemes. This can be seen on the test equation $\frac{\diff y}{\diff t} = \lambda y$, where Euler implicit schemes are always stable while Euler explicit schemes are not. Interpreting residual neural networks as numerical schemes, one can apply such schemes and gain theoretical insights on the properties of neural networks. This has already been done in various forms in \citet{haber2017stable,chen2018neural}, but not applied to forecasting. Moreover, these works use either the stability of the underlying continuous equation or the stability of the numerical scheme on the test equation and its derivatives, which is not the stability of the numerical scheme on the studied equation. Since a network is discrete, the latter is the most relevant.

We therefore use the stability in norm of schemes, as defined in \ref{def:def_stab_in_norm}. In deep learning (DL), this definition has only been applied to image classification problems \citep{zhang2020forward}. To the best of our knowledge, this work is the first attempt to forecast PDEs with neural networks using stability as studied in the numerical analysis community.

Using implicit schemes in DL has already been done in different contexts. The earliest line of works tackles image problems, with \citet{haber2019imexnet} designing semi-implicit ResNets and \citet{li2020implicit,shen2020implicit, reshniak2021robust} designing different implicit ResNets.
The second line of works focuses on dynamical problems. In this way, \citet{nonnenmacher2021learning} designed linear implicit layers, which learn and solve linear systems, and \citet{horie2022physics} used an implicit scheme as part of an improvement of graph neural network solvers to improve forecasting generalization with different boundary condition shapes. Tackling different types of problems, none of these methods guarantees the forecast stability. For our analysis, we restrict ourselves to ResNet-type networks, \ie, networks with residual connections. We introduce hard constraints on the weights of the network and predict within the latent space of our network. Hence, by modifying the classic implicit ResNet architecture, our method can forecast dynamical system at long range without diverging. We apply these theoretical constraints in our architecture
to two 1D transport problems: the \textit{Advection equation} and \textit{Burgers' equation}. 

To better investigate networks stability, we perform our experiments under the following challenging setting: for training, we only give to the networks the data from $t=0$ to a small time $t=\Delta t$, and consider the results in forecasting in the long time range at $t=N \cdot \Delta t$, with $N\gg1$. Note that our setting is harder that the conventional setting presented for e.g. in \cite{brandstetter2022message}. Indeed, we only use changes between a single time step for training.

\section{Method}

To guarantee structural forecasting stability, we take inspiration from implicit schemes. We focus our study on an implicit ResNet with a ReLU activation function. 
 In our approach, an equation is solved at each layer, namely $x_{n+1} := x_n + R_n(x_{n+1})$ with $x$ in $\mathbb{R}^M$ and  $n$ in $\mathbb{N}$ and $R_n(x)=\text{ReLU}(W_n x+b_n)$ where $W_n$ is upper triangular. The latter constraint is motivated below.

\subsection{Implicit ResNet stability analysis} \label{sec:stab_analysis}
To ensure that our proposed architecture is well-defined, we need to solve $x = x_n + R_n(x)$. This equation has a solution, as proven in \citet{el2019implicit} and detailed in Appendix \ref{sec:app_fixed_pt_sol}. We can then study its stability. The recursion defining $(x_n)_{n \in \mathbb{N}}$ reads as an implicit Euler scheme with a step size of 1. As described in the introduction, an implicit scheme is usually more stable than an explicit one. We first recall the definition of stability for schemes. This property ensures that our architecture has an auto-regressive stability.
\begin{definition}[Stability in norm]\label{def:def_stab_in_norm}
A scheme $(x_n)_{n \in \mathbb{N}}$ of dimension $M$ is stable in norm $L^p$ if there exists $C(T)$ independent of 
the time discretization step $\Delta t$ such that:
\begin{equation*}
     \forall ~ x_0 \in \mathbb{R}^M, ~ n \ge 0; ~ n \Delta t \leq T, \text{    } \norm{x_n}_{p} \leq C(T) \norm{x_0}_{p} \text{     .}  
\end{equation*}
\end{definition}
This general definition of stability in norm ensures that a scheme does not amplify errors. This definition is equivalent to several others in the numerical analysis community.
\newline
Suppose that the spectrum of $W_n$ is contained in $[-1,0[$ for every integer $n$, we can assert that $(x_n)_{n \in \mathbb{N}}$ is well-defined, using theorem \ref{theorem:root_existence}. The proof of the stability of our Implicit ResNet network is then by induction on the dimension and is given in appendix \ref{sec:proof}:

\begin{theorem}[Stability theorem] \label{theorem:stab_in_norm}
If the elements of the diagonal of $W_n$ are in $[-1,0)$ for every integer $n$, then $(x_n)_{n \in \mathbb{N}}$ is stable in norm $L^p$.
\end{theorem}
This theorem leads to hard constraints on the weights of the network. 
\subsection{Implementation}
To validate practically our theoretical results, we choose a setting that highlights stability issues. We then test our implementation of an implicit ResNet. In order to respect the assumptions of theorem \ref{theorem:stab_in_norm}, we forecast the dynamics in the latent space, as detailed below.
\paragraph{Setting} \label{sec:setting_main}
We first learn the trajectory at a given small time step $\Delta t$. We only give data $t=0$ to $t=\Delta t$ for the training. 
We then forecast in long-term, at $N \cdot \Delta t$ with $N\gg1$. This very restricted setting allows us to see how the network will react in forecasting with changes in the distribution and error accumulation. Usually neural network forecasting methods use data from $t=0$ to $T=L\cdot\Delta t$ for the training which allows to use different tricks to stabilize the network, such as predicting multiple time steps at the same time. However, in this work, we want to analyze how the network behaves without any trick that can slow down divergence. Indeed, the tricks used in the other settings do not actually guarantee stability of the network. The training is performed with a mean-squared error (MSE) loss.

\paragraph{Implicit neural network architecture} \label{sec:method_def}
To implement a neural network from Theorem \ref{theorem:stab_in_norm}, we use the following architecture; $z_{\Delta t} = f_{dec} \circ f_{res}^K \circ ... \circ f_{res}^1 \circ f_{enc} (z_0)$, with $f_{res}^k(x) = x + \text{ReLU}(W_{k-1} \cdot f_{res}^k(x) + b_{k-1})$.
The encoder and decoder are linear, and the encoder projects the data to a smaller dimension $M$. The full architecture is illustrated in Figure \ref{fig:implicit_nn_archi}.
The specificity of our architecture is that the residual blocks are connected with an implicit Euler scheme iteration. To do so, we use a differentiable root-finding algorithm \citep{kasim2020xi}. More precisely, we use the first Broyden method \cite{van2005limited} to solve the root-finding problem. It is a quasi-Newton method. This helped getting better results compared to other algorithms.

\paragraph{Latent space forecasting}
As in \citet{wiewel2019latent,kim2019deep}, the forecast can be done within the latent space of the network; $z_{N\cdot\Delta t} = f_{dec} \circ f_{block}  \circ ...  \circ f_{block} \circ f_{enc} (z_0)$, with $f_{block} = f_{res}^K \circ ... \circ f_{res}^1$. To predict at time $N \cdot \Delta t$ from time $t=0$, we apply $N$ times the residual blocks. It is illustrated in Figure \ref{fig:latent_space_forecasting}. This propagation allows our network to respect the assumptions of theorem \ref{theorem:stab_in_norm} and thus be stable.

\section{Experiments}
We evaluate the performance of our method on the \textit{Advection equation} and \textit{Burgers' equation}.
\paragraph{Datasets}
\textit{Advection equation}. We construct our first dataset with the following 1-D linear PDE, 
$\frac{\partial\psi}{\partial t} = -\frac{1}{4} \frac{\partial \psi}{\partial z}, z \in (0, 2\pi), \, t \in \mathbb{R}^{+}$ and $\psi(z,0) = f_0(z), z \in (0,2\pi)$.

\textit{Burgers' equation}. In the second dataset, we consider the following 1-D non-linear PDE, 
\newline
$\frac{\partial\psi}{\partial t} = -\frac{1}{2} \frac{\partial \psi^2}{\partial z} + \frac{\partial^2 \psi^2}{\partial z}, z \in (0, 2\pi), \, t \in (0,1]$ and $\psi(z,0) = f_0(z), z \in (0,2\pi)$. 
\vspace{0.2cm}
\newline
Both equations have periodic boundary conditions. We approximate the mapping from the initial condition $f_0$ to the solution at a given discretization time $\Delta t$, \ie $u_0 \mapsto u(\cdot, \Delta t)$. We then forecast to longer time ranges. $\Delta t$ is set to 1 for the \textit{Advection equation} with a grid of 100 points and 0.0005 for \textit{Burgers' equation} with a grid of 128 points.

\begin{figure}[ht!]
    \begin{subfigure}{0.5\columnwidth}
        \centering
        \includegraphics[width = 0.8\linewidth]{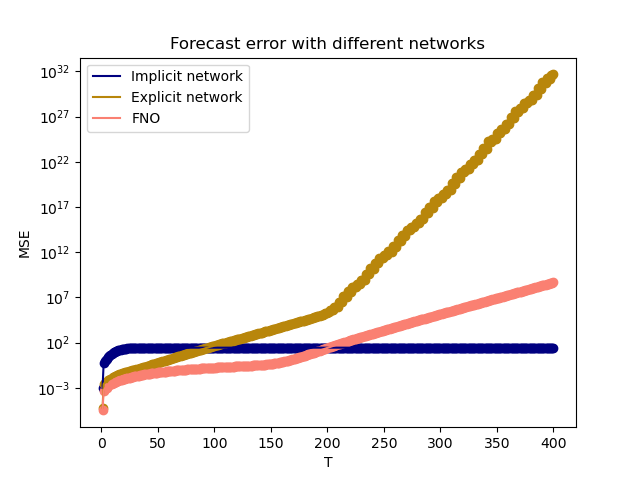}
    \end{subfigure}
    \begin{subfigure}{0.5\columnwidth}
        \centering
        \includegraphics[width = 0.8\linewidth]{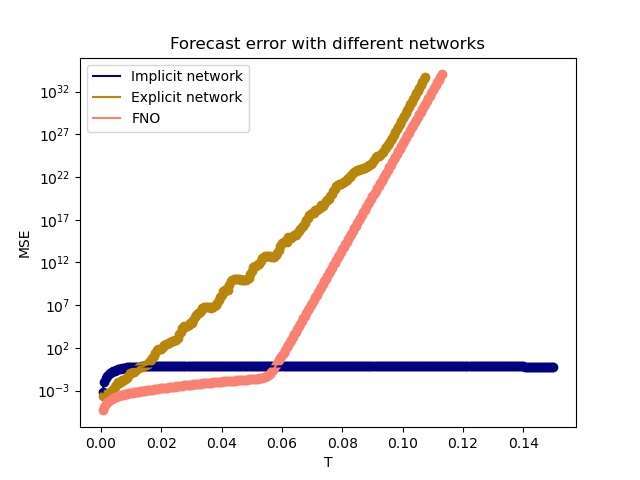}
        
    \end{subfigure}
    \caption{Forecast error for different neural network architectures for the \textit{Advection equation} (left) and \textit{Burger's equation} (right).}
    \label{fig:mse_different_methods}
\end{figure}

\paragraph{Baseline methods} \label{sec:baseline_prez}
We compare our Implicit ResNet with respect to an Explicit ResNet with ReLU activation function and a Fourier Neural Operator (FNO) \citep{li2020fourier}. 
We have also implemented two Explicit ResNet, with a tanh activation function and with batch normalization. We design the Explicit ResNet in the same way as our implicit version, with $K$ layers of residual blocks that are linked by $x_{n+1} = x_n + R_n(x_{n})$.
Traditional methods forecast by using the output of the network at time $t$ to predict the dynamics at time $t+\Delta t$. So to predict at time $N \cdot \Delta t$ from time $t=0$, the baseline networks are iteratively applied $N$ times, as illustrated in Figure \ref{fig:tradi_auto_reg_forecasting}.
\paragraph{Results}
Prediction errors are reported in Table \ref{tab:main_res} for the \textit{Advection equation} and \textit{Burgers' equation}. We also show the error according to the forecast time in Figure \ref{fig:mse_different_methods}.

We found that traditional deep learning methods diverge with the forecast time. They reach a MSE of more than $10^8$ for the \textit{Advection equation} and go to infinity for \textit{Burgers' equation}, respectively at time $400$ and $0.15$. Moreover, we see in Figure \ref{fig:mse_different_methods} that their curve in time is convex, so the increase in error is accelerating over time. We also found that our proposed Implicit ResNet presents better results by several orders of magnitude for both datasets. Moreover, we can see in Figure \ref{fig:mse_different_methods} that its curve in time is reaching a stable plateau, as expected from our theorem \ref{theorem:stab_in_norm}.

As for the training, traditional deep learning methods manage to learn very well the dynamics at $t=\Delta t$, with two orders of magnitude better than our Implicit ResNet. However, the latter still manages to learn well the dynamics with a MSE of $10^{-2}$ for the \textit{Advection equation} and $10^{-3}$ for \textit{Burgers' equation}. This difference in training can mainly be explained by the longer training time of the Implicit ResNet, which made us take a smaller number of epochs for this network (1250 against 2500).
\begin{table}[ht!]
    \caption{Results of our approach compared to baselines on the \textit{Advection equation} and \textit{Burgers' equation}. We calculate the means and standard deviations of MSE for each model based on 5 runs with different seeds. The mid-range time is 40 for the \textit{Advection equation} and 0.075 for \textit{Burgers'}  and the long range time is respectively 400 and 0.15. Recall that $\Delta t_{adv} = 1$ and  $\Delta t_{bur} = 0.0005$.
    } 
    \centering
    \resizebox{\textwidth}{!}{  
    \begin{tabular}{lllllll}
\toprule
        &  &\bf Model &\bf \makecell[l]{Train Error \\ ($\times 10^{-4}$)}&\bf \makecell[l]{Test Error \\ ($\times 10^{-4}$)} &\bf \makecell[l]{Forecast error at mid-range \\ $T_{adv} = 40 \cdot \Delta t_{adv}$ \\ $T_{bur} = 150 \cdot \Delta t_{bur}$} &\bf \makecell[l]{Forecast error at long-range \\ $T_{adv} = 400 \cdot \Delta t_{adv}$ \\ $T_{bur} = 300 \cdot \Delta t_{bur}$} \\
        \midrule
      \multirow{3}{*}{\rotatebox[origin=c]{45}{\bf \textit{Advection}}} &\multirow{3}{*}{\rotatebox[origin=c]{15}}
       & Explicit Res Net & \textbf{0.03 ± 0.01} & \textbf{0.09 ± 0.07}  & 0.25 ± 0.33 & $4.7 \cdot 10^{31}$ ± $1.0 \cdot 10^{32}$ \\
       & & FNO  & \textbf{0.04 ± 0.01} & \textbf{0.1 ± 0.08} & \textbf{0.03 ± 0.04} & $4.7 \cdot 10^{8}$ ± $1.0 \cdot 10^{9}$ \\
       & & Implicit ResNet (Ours)  & 14.0 ± 9.0 & 25.0 ± 27.0 & 27.4 ± 24.0 & \textbf{27.5 ± 24.2} \\
        \midrule
        \multirow{3}{*}{\rotatebox[origin=c]{45}{\bf \textit{Burgers'}}} &\multirow{3}{*}{\rotatebox[origin=c]{15}}
       & Explicit Res Net & 0.17 ± 0.03 & 0.90 ± 0.38 & $2.77 \cdot 10^{19}$ ± $6.2 \cdot 10^{19}$  & $+ \infty$  \\
       & & FNO  & \textbf{0.02 ± 0.002} & \textbf{0.03 ± 0.006} & $5.31 \cdot 10^{10}$ ± $11.2 \cdot 10^{10}$ & $+ \infty$ \\
       & & Implicit ResNet (Ours)  & 4.90 ±  0.64 & 7.91 ± 0.30 & \textbf{0.67 ± 0.43} & \textbf{0.66 ± 0.44}\\
    \bottomrule
    \end{tabular}
    }
    \label{tab:main_res}
\end{table}

\paragraph{Discussion}
Figure \ref{fig:mse_different_methods} demonstrates the main benefits of our constrained implicit neural network. Our network is stable whereas the other methods diverge in time. However, although being stable and far better than the baselines, it does not manage to forecast accurately the long-term dynamics. This is further confirmed by Table \ref{tab:rel_main_res}, which shows high relative errors. Said otherwise, when stability is guaranteed, convergence is not. We can also note that constraining the weights makes our network harder to train, but guarantees structural forecasting stability.
\section{Conclusion}
In this work, we studied the challenging task of long-term forecasting in dynamical systems. To do so, we developed a theoretical framework to analyze the stability of deep learning methods for forecasting dynamical systems. We then designed a constrained implicit neural network out of this framework. To the best of our knowledge, this is the first work that proposes to study deep learning architectures for forecasting dynamical systems from a numerical schema standpoint. We showed improved results with respect to deep learning baselines for two transport PDEs. 

This work opens new perspectives to study neural networks forecasting stability from a numerical schema standpoint, thus offering more robust architectures. However, this analysis still needs improvements. Even though it ensures forecasting stability of our proposed network, it does not guarantee good convergence properties. 
We believe that developing this line of research could help overcome these challenges, and provide more robust architectures for forecasting dynamical systems in long time range. 
\newpage

\section*{Acknowledgments}
We thank Yuan Yin and Etienne Le Naour for helpful insights and discussions on this project.

\bibliographystyle{abbrvnat}
\bibliography{bibtex}

\section{Details on Implicit ResNet stability analysis}

\subsection{Fixed point solution existence} \label{sec:app_fixed_pt_sol}
We first define the Perron–Frobenius eigenvalue, before stating the root existence, which uses this eigenvalue. Let $M$ be a non-negative square matrix, i.e. with non-negative entries.
\begin{theorem}[Perron–Frobenius theorem]
$M$ admits a real eigenvalue that is larger than the modulus of any other eigenvalue.
\end{theorem}
This non negative eigenvalue is called the Perron–Frobenius (PF) eigenvalue and is denoted $\lambda_{pf}(M)$.
\begin{theorem}[Root existence] \label{theorem:root_existence}
For an integer $n$, given that ReLU is non-expansive, if \newline $\lambda_{pf}(|W_n|)<1$, then $x$ defined as $x = x_n + R_n(x)$ exists.
\end{theorem}
The proof is available in theorem 2.2 of \citet{el2019implicit}. They show that the solution can be obtained using a fixed-point iteration. However they do not offer any analytical solution.

\subsection{Definitions and notations} \label{sec:def_and_not}
Let $(\alpha_{n}^{(m_1, \, m_2)})_{m_1, \, m_2 \in \in [1:M]^2}$ be the strict upper entries 
of  $W_n$ and $(b_{n}^{(m)})_{m \in [1:M]}$ the entries of $b_n$. We suppose that $(\alpha_{n}^{(m_1, \, m_2)})_{n \in \mathbb{N}}$ and $(b_{n}^{(m)})_{n \in \mathbb{N}}$ are bounded. Let $Q:=\max_{m_1 \in [|0,M-1|], \, m_2 \in [|m_1+1, m|]}(\max_{n \in \mathbb{N}}( |\alpha_{n}^{(m_1, \, m_2)})|))$ and $B:=\max_{n \in \mathbb{N}, \, m \in [|1,M|]}(b_{n}^{(m)})$. Let $(-\lambda_{n}^{(m)})_{m \in [1:M]}$ be the entries of the diagonal of $W_n$, and $P:=\min_{n \in \mathbb{N}, \, m \in \llbracket 1,M\rrbracket} (\lambda_{n}^{(m)}))$. $P$ is by hypothesis finite and positive.
\newline
For an integer $n$ and $x_n \in \mathbb{R}^M$, we will denote by $x_n^{(m)}$ the $n$\ts{th} iteration of the $m$\ts{th} dimension of the sequence $(x_n)$. For $m$ in $[|1, M|]$, let $S_m:=\max_{j \in [|1,m|], \, k \in \mathbb{N}}(x_k^{(j)})$ and $S_0 = 0$. 

\subsection{Explicit expression of \texorpdfstring{$v_n$}{TEXT}} \label{sec:solving_vn}
The definitions and notations detailed in section \ref{sec:def_and_not} are used throughout this section.
\begin{definition} \label{sec:def_vn_m} 
We define, for an integer $n$ and $m$ in $[|0, M-1|]$, $v_n^{(m)}$ by the recursion:
\begin{equation}
    v_{n+1}^{(m)}:=\frac{v_n^{(m)} + \sum_{j=1}^{m-1} \alpha_{n}^{((m-1), \, j)} x_{n+1}^{(j)}+b_{n}^{(m)}}{1 +\lambda_{n+1}^{(m)}} . 
\end{equation}
\end{definition}

\begin{lemma}[Explicit expression of $v_n^{(m)}$] \label{lemma:expression_vn}
For an integer $n$ and $m$ in $[|1, M|]$, an explicit expression of $v_n^{(m)}$ is given by:
\begin{equation*}
    v_{n}^{(m)} = x_{0}^{(m)} \prod_{k=1}^{n} \frac{1}{1 +\lambda_{k}^{(m)}} +  \sum_{k=1}^{n} (\prod_{l=k}^{n} \frac{1}{1 +\lambda_{l}^{(m)}} \sum_{j=1}^{m-1} \alpha_{k-1}^{((m-1), \, j)} x_{k}^{(j)}) + \sum_{k=1}^{n} (\prod_{l=k}^{n} \frac{1}{1 +\lambda_{l}^{(m)}} b_{k-1}^{(m)}) \label{eq:expli_expre_vn} .
\end{equation*}
\end{lemma}
\begin{proof}
In order to obtain an explicit expression of $v_n^{(m)}$, we write out all the terms of $v_n^{(m)}$. Let $i$ be an integer in $[0, n]$. We then multiply each term by $\prod_{k=2}^{i+1} \frac{1}{1 +\lambda_{k}^{(m)}}$:
\begin{equation}
    \prod_{k=2}^{i+1} \frac{1}{1 +\lambda_{k}^{(m)}}(v_{n+1 - i}^{(m)} - \frac{1}{1+\lambda_{n+1-i}^{(m)}} x_{n-i}^{(m)}) = ( \frac{\sum_{j=1}^{m-1} \alpha_{n-i}^{((m-1), \, j)} x_{n+1-i}^{(j)} + b_{n-i}^{(m)}}{1 +\lambda_{n+1-i}^{(m)}})\prod_{k=2}^{i+1} \frac{1}{1 +\lambda_{k}^{(m)}} . \label{eq:solve_v_n_i}
\end{equation}
We thus obtain a telescoping sum by adding equations Eq. \eqref{eq:solve_v_n_i} for every $i$ in $[0, n]$.
\end{proof}

\subsection{Proof of theorem \ref{theorem:stab_in_norm}} \label{sec:proof}
The definitions and notations detailed in section \ref{sec:def_and_not} are used throughout this section.
\begin{proof}
We will prove that, for $m$ in $[|1, M|]$,  $(x_{n}^{(m)})_{n \in \mathbb{N}}$ is bounded. The proof is by induction on $m$.

For the base case $m=1$, let $n$ be an integer. We will show that $(x_{n}^{(1)})_{n \in \mathbb{N}}$ is bounded.

    It is easily seen that:
    \begin{equation*}
            x_{n+1}^{(1)} =   \left\{
        \begin{array}{ll}
        x_{n}^{(1)} &\text{, if  } -\lambda_{n+1}^{(1)} x_{n}^{(1)}+b_{n}^{(1)} \leq 0 \\
         \frac{1}{1 +\lambda_{n+1}^{(1)}}(x_{n}^{(1)}+b_{n}^{(1)}) &\text{, else.}
        \end{array}\right.
    \end{equation*} 
        Let $u_n^{(1)}:=x_{0}^{(1)}$ and $v_{n+1}^{(1)}:=\frac{v_n^{[1)} + b_{n}^{(1)}}{(1 +\lambda_{n+1}^{(1)})}$. We then have $\min(u_n^{(1)}, v_n^{(1)}) \leq x_{n}^{(1)} \leq \max(u_n^{(1)}, v_n^{(1)})$.
        \newline
        Using lemma \ref{lemma:expression_vn}, $v_n^{(1)}$ may be written as:
    \begin{equation}
        v_{n+1} = x_{0}^{(1)} \prod_{k=1}^{n+1} \frac{1}{(1 +\lambda_{k}^{(1)})} + \sum_{k=1}^{n+1} (\prod_{l=k}^{n+1} \frac{1}{(1 +\lambda_{l}^{(1)})} b_{k-1}^{(1)}) .\label{eq:global_expr_vn_1}
    \end{equation}
    We can then bound the second term of the right-hand side of Eq. \eqref{eq:global_expr_vn_1}:
    \begin{align}
        |\sum_{k=1}^{n+1} (\prod_{l=k}^{n+1} \frac{1}{(1 +\lambda_{l}^{(1)})}b_{k-1}^{(1)})| &\leq  
        B \sum_{k=1}^{n+1} (\frac{1}{(1 +P)^{n+1-k}} \notag \\
        &\leq B \frac{(1 +P)^{n+1} - (1+P)}{P(1 +P)^{n+1}} . \label{eq:bound_right_side_vn_1}
    \end{align}
    Combining Eq. \eqref{eq:global_expr_vn_1} and \eqref{eq:bound_right_side_vn_1}, we can assert that $v_n^{(1)}$ is bounded.
    \newline 
    Since $\min(x_0^{(1)}, v_n^{(1)}) \leq x_{n}^{(1)} \leq \max(x_0^{(1)}, v_n^{(1)})$, we can conclude that $(x_{n}^{(1)})_{n \in \mathbb{N}}$ is bounded.
    \newline
    Suppose $\forall ~ j \in [|0,m|], (x_{n}^{(j)})_{n \in \mathbb{N}}$ bounded, we will now prove that $(x_{n}^{(m+1)})_{n \in \mathbb{N}}$ is bounded.
    \newline
    We first solve Eq. \eqref{eq:eq_ND_x} to find an expression of $x_{n+1}^{(m+1)}$.
    \begin{equation}
        X = x_{n}^{(m+1)} + \max(0,-\lambda_{n+1}^{(m+1)} X + \sum_{j=1}^{m} \alpha_{n}^{(m, \, j)} x_{n+1}^{(j)} + b_{n}^{(m+1)}) . \label{eq:eq_ND_x}
    \end{equation}

    Deriving both cases, we obtain:
    \begin{equation*}
        x_{n+1}^{(m+1)} = \left\{
        \begin{array}{ll}
            x_{n}^{(m+1)} &\text{, if  } -\lambda_{n+1}^{(m+1)}x_{n}^{(m+1)}+\sum_{j=1}^{m} \alpha_{n}^{(m, \, j)} x_{n+1}^{(j)} + b_{n}^{(m+1)} \leq 0 \\
            \frac{x_{n}^{(m+1)}+\sum_{j=1}^{m} \alpha_{n}^{(m, \, j)} x_{n+1}^{(j)}+b_{n}^{(m+1)}}{(1 +\lambda_{n+1}^{(m+1)})} &\text{, else.}
        \end{array}\right. 
    \end{equation*}
    The proof is left to the reader.
    \newline
    Let $u_n^{(m+1)}:=x_{0}^{(m+1)}$ and $v_{n+1}^{(m+1)}:=\frac{v_n^{(m+1)} + \sum_{j=1}^{m} \alpha_{n}^{(m, \, j)} x_{n+1}^{(j)}+b_{n}^{(m+1)}}{(1 +\lambda_{n+1}^{(m+1)})}$. We then have that \[\min(u_n^{(m+1)}, v_n^{(m+1)}) \leq x_{n}^{(m+1)} \leq \max(u_n^{(m+1)}, v_n^{(m+1)}).\]
    \newline
    Using lemma \ref{lemma:expression_vn}, $v_n^{(m+1)}$ may be written as:
    \begin{equation*}
        v_{n+1}^{(m+1)} = x_{0}^{(m+1)} \prod_{k=1}^{n+1} \frac{1}{1 +\lambda_{k}^{(m+1)}} +  \sum_{k=1}^{n+1} (\prod_{l=k}^{n+1} \frac{1}{1 +\lambda_{l}^{(m+1)}} \sum_{j=1}^{m} \alpha_{k-1}^{(m, \, j)} x_{k}^{(j)}) + \sum_{k=1}^{n+1} (\prod_{l=k}^{n+1} \frac{1}{1 +\lambda_{l}^{(m+1)}} b_{k-1}^{(m+1)}) .
    \end{equation*}
    It is easily seen that the first and third terms of $v_n^{(m+1)}$ are bounded. We still wish to bound the second term of $v_n^{(m+1)}$. Using the induction hypothesis, $S_m$ is finite. We can then bound the second term of $v_n^{(m+1)}$:
    \begin{align}
        |\sum_{k=1}^{n+1} (\prod_{l=k}^{n+1} \frac{1}{1 +\lambda_{l}^{(m+1)}} \sum_{j=1}^{m} \alpha_{k-1}^{(m, \, j)} x_{k}^{(j)})| &\leq  
        |\sum_{k=1}^{n+1} (\frac{1}{(1 +P)^{n+1-k}} \sum_{j=1}^{m} \alpha_{k-1}^{(m, \, j)} x_{k}^{(j)})| \notag \\
        &\leq | \sum_{k=1}^{n+1} \frac{1}{(1 +P)^{n+1-k}} m Q S_m| \notag \\
        &\leq m Q S_m  \frac{1}{(1 +P)^{n+1}} \sum_{k=1}^{n+1} ((1+P)^{k} \notag \\
        &\leq m Q S_m \frac{(1 +P)^{n+2} - (1+P)}{P(1 +P)^{n+1}} . \label{eq:bound_thrid_term_vn}
    \end{align}
    Since Eq. \eqref{eq:bound_thrid_term_vn} shows that the second term of $v_n^{(m+1)}$ is bounded, $v_n^{(m+1)}$ is bounded, hence we can conclude that $x_{n}^{(m+1)}$ is bounded.
    \newline
    Since both the base case and the induction step have been proved as true, by mathematical induction for every $m$ in $[|1, M|]$, $(x_{n}^{(m)})_{n \in \mathbb{N}}$ is bounded. Hence $x_n=(x_{n}^{(1)}, ..., x_{n}^{(M)})$ is bounded.
\end{proof}

\section{Details on the implementation}
\subsection{Implicit neural network architecture} \label{sec:appendix_imp_resnet}
\begin{figure}[h!]
    \centering
    \includegraphics[width = 0.9\linewidth]{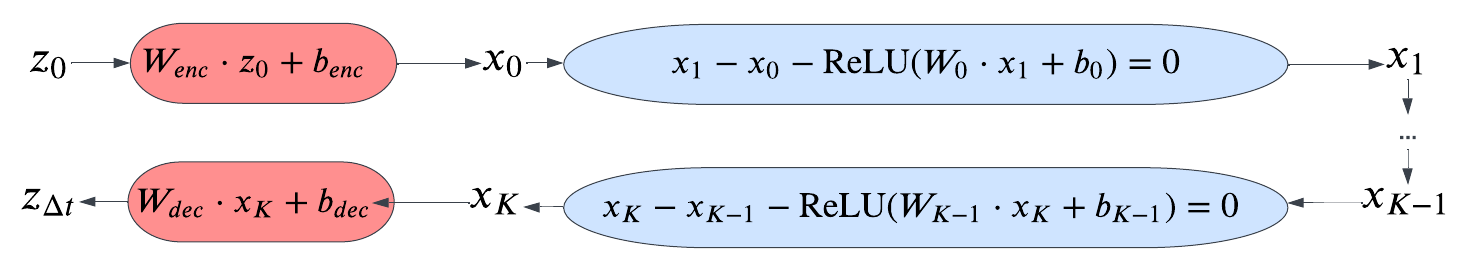}
    \caption{Implicit neural network architecture with K residual blocks.}
    \label{fig:implicit_nn_archi}
\end{figure}

Our implicit neural network is using Rectified linear unit (ReLU) activation functions, as can be seen in Figure \ref{fig:implicit_nn_archi}. 
\begin{definition}[ReLU]\label{def:relu}
A rectified linear unit (ReLU) function is defined component-wise to a vector by $\forall ~ x \in \mathbb{R}, \text{ReLU}(x) = \max(0,x)$. 
\end{definition}
It is one of the most common activation functions used in Deep Learning.

In order to constrain our network, we use upper triangular weights $W_n$. At each training epoch, we constrain the diagonal values to be between -1 and 0 after gradient descent. We choose a minimal value of 0.01, to ensure that the theorem hypothesis are respected. For values below -1, we set them to -1 and for values above 0.01, we set them to 0.01.

 \subsection{Forecasting settings}
\begin{figure}[h!]
    \centering
    \includegraphics[width = 0.9\linewidth]{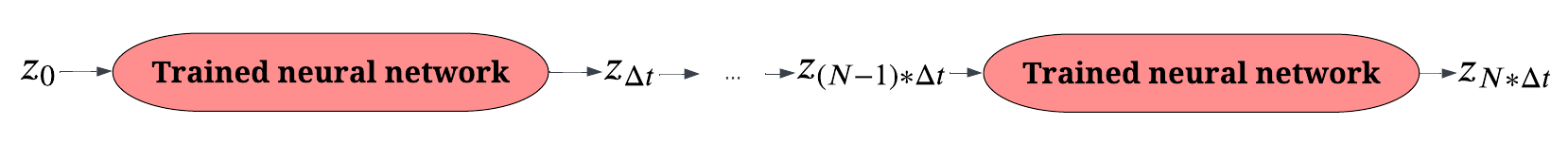}
    \caption{Traditional auto-regressive forecasting.}
    \label{fig:tradi_auto_reg_forecasting}
\end{figure}

\begin{figure}[h!]
    \centering
    \includegraphics[width = 0.9\linewidth]{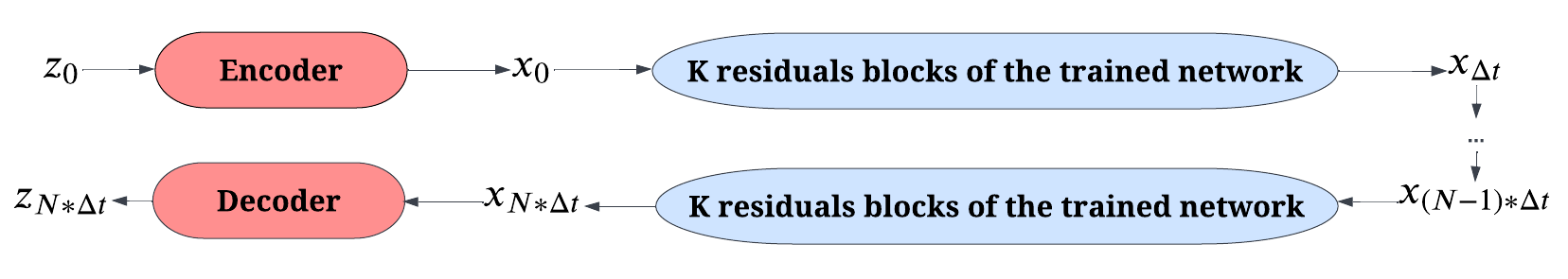}
    \caption{Latent-space auto-regressive forecasting.}
    \label{fig:latent_space_forecasting}
\end{figure}

 As described in section \ref{sec:setting_main}, traditional methods forecast by using the output of the network at time $t$ to predict the dynamics at time $t+\Delta t$. Figure \ref{fig:tradi_auto_reg_forecasting} illustrates this setting. However, the forecast can also be done within the latent space of the network. Figure \ref{fig:latent_space_forecasting} illustrates this different setting.

\section{Details on experiments}

\subsection{Baseline Methods}
In addition to an Explicit ResNet with ReLU activation function and a FNO, we have two variants for the explicit ResNet method.
\begin{itemize}
    \item an explicit ResNet tanh, with $R_n(x) = \tanh(W_n x+b_n)$
    \item an explicit ResNet BN, with a ReLU activation function, and batch normalization at each hidden layer, to control the norm inside the network.
\end{itemize}
 
\subsection{On training} \label{sec:on_training}
The training details for each architecture on both equations are presented in the appendix in Table \ref{tab:hyper_param_networks}. For the \textit{Advection equation}, 350 examples were generated for train, 150 other ones for validation/test and 50 for forecasting tests. For \textit{Burgers' equation}, 120 examples were generated for train, 30 other ones for validation/test and 50 for forecasting tests. Our experiments led to a few main training remarks.
\paragraph{Initialization} The networks are not really sensitive to the initialization. The only pitfall is to initialize with high values. Then the network doesn't manage to converge as well as it could have. We initialize all networks with Xavier initialization with a gain of 1.
\paragraph{Learning rate scheduling} We used learning rate scheduling. It improves performance by a factor of 100. It is crucial to use it for our problems, and to choose carefully its parameter. We found that a linear scheduling with carefully chosen decay and step size works well. A special attention needs to be placed on the initial learning rate as well.
\paragraph{FNO architecture} For the FNO network, we chose 12 modes a width of 32 for the \textit{Advection equation} and 16 modes and a width of 64 for \textit{Burgers' equation}, as was done in the original article.

\subsection{Training parameters}
The training remarks detailed previously in section \ref{sec:on_training} led to the choices showed in Table \ref{tab:hyper_param_networks}.
\begin{table}[H]
    \caption{Hyper-parameter choice for each architecture on the \textit{Advection equation} and \textit{Burgers' equation}.
    } 
    \centering
    \resizebox{\textwidth}{!}{  
    \begin{tabular}{llllllll}
\toprule
        &  &\bf Model &\bf \makecell[l]{Xavier gain}&\bf \makecell[l]{Initial learning rate} &\bf \makecell[l]{Decay} &\bf \makecell[l]{Step size} &\bf \makecell[l]{Epochs} \\
        \midrule
      \multirow{6}{*}{\rotatebox[origin=c]{45}{\bf \textit{Advection}}} &\multirow{3}{*}{\rotatebox[origin=c]{15}}
       & Explicit Res Net & 1 & 0.05 & 0.95 & 10 & 2500  \\
       &  & Explicit Res Net BN & 1 & 0.05 & 0.98 & 10 & 2500\\
       &  & Explicit Res Tanh & 1 & 0.05 & 0.95 & 10 & 2500\\
       & & FNO  & 1 & 0.005 & 0.98 & 10 & 2500 \\
       & & Implicit ResNet (Ours)  & 1 & 0.01 & 0.9 & 10 & 1250 \\
        \midrule
        \multirow{6}{*}{\rotatebox[origin=c]{45}{\bf \textit{Burgers'}}} &\multirow{3}{*}{\rotatebox[origin=c]{15}}
       & Explicit Res Net & 1 & 0.05 & 0.95 & 10 & 2500 \\
       &  & Explicit Res Net BN & 1 & 0.05 & 0.98 & 10 & 2500 \\
       &  & Explicit Res Tanh & 1 & 0.05 & 0.95 & 10 & 2500\\
       & & FNO  & 1 & 0.005 & 0.96 & 10 & 2500 \\
       & & Implicit ResNet (Ours)  & 1 & 0.01 & 0.98 & 10 & 1250 \\
 \bottomrule
    \end{tabular}
    }
    \label{tab:hyper_param_networks}
\end{table}

\subsection{Ablation study}
In order to better investigate this task, we conducted experiments with additional architectures. The results are shown in Table \ref{tab:ablation_study}.
\begin{table}[H]
    \caption{Ablation study for the \textit{Advection equation} and \textit{Burgers' equation}.
    } 
    \centering
    \resizebox{\textwidth}{!}{  
    \begin{tabular}{lllllll}
\toprule
        &  &\bf Model &\bf \makecell[l]{Train Error \\ ($\times 10^{-4}$)}&\bf \makecell[l]{Test Error \\ ($\times 10^{-4}$)} &\bf \makecell[l]{Forecast error at mid-range \\ $T_{adv} = 40 \cdot \Delta t_{adv}$ \\ $T_{bur} = 150 \cdot \Delta t_{bur}$} &\bf \makecell[l]{Forecast error at long-range \\ $T_{adv} = 400 \cdot \Delta t_{adv}$ \\ $T_{bur} = 300 \cdot \Delta t_{bur}$} \\
        \midrule
      \multirow{6}{*}{\rotatebox[origin=c]{45}{\bf \textit{Advection}}} &\multirow{3}{*}{\rotatebox[origin=c]{15}}
       & Explicit Res Net & \textbf{0.03 ± 0.01} & \textbf{0.09 ± 0.07}  & 0.25 ± 0.33 & $4.7 \cdot 10^{31}$ ± $1.0 \cdot 10^{32}$ \\
       &  & Explicit Res Net BN & 1.01 ± 0.32 & 317 ± 20  & $1.2 \cdot 10^{24}$  ± $2.8 \cdot 10^{24}$  & $+ \infty$ \\
       &  & Explicit Res Tanh & 0.98 ± 0.1 & 14.0 ± 4.0  & 31.7  ± 70.5 & $+ \infty$  \\ 
       & & FNO  & \textbf{0.04 ± 0.01} & \textbf{0.1 ± 0.08} & \textbf{0.03 ± 0.04} & $4.7 \cdot 10^{8}$ ± $1.0 \cdot 10^{9}$ \\
       & & Implicit ResNet (Ours)  & 14.0 ± 9.0 & 25.0 ± 27.0 & 27.4 ± 24 & \textbf{27.5 ± 24.2} \\
        \midrule
        \multirow{6}{*}{\rotatebox[origin=c]{45}{\bf \textit{Burgers'}}} &\multirow{3}{*}{\rotatebox[origin=c]{15}}
       & Explicit Res Net & 0.17 ± 0.03 & 0.90 ± 0.38 & $2.77 \cdot 10^{19}$ ± $6.2 \cdot 10^{19}$  & $+ \infty$  \\
       &  & Explicit Res Net BN & 0.51 ± 0.13 & 51.63 ± 34.89 & $+ \infty$ & $+ \infty$ \\
       &  & Explicit Res Tanh & 0.84 ± 0.22 & 44.67 ± 11.58 & $+ \infty$ & $+ \infty$ \\
       & & FNO  & \textbf{0.02 ± 0.002} & \textbf{0.03 ± 0.006} & $5.31 \cdot 10^{10}$ ± $11.2 \cdot 10^{10}$ & $+ \infty$ \\
       & & Implicit ResNet (Ours)  & 4.90 ±  0.64 & 7.91 ± 0.30 & \textbf{0.67 ± 0.43} & \textbf{0.66 ± 0.44}\\
 \bottomrule
    \end{tabular}
    }
    \label{tab:ablation_study}
\end{table}

\begin{table}[h!]
    \caption{Results of our approach compared to baselines on the \textit{Advection equation} and \textit{Burgers' equation}. We calculate the means and standard deviations of relative error for each model based on 5 runs with different seeds. The mid-range time is 40 for the \textit{Advection equation} and 0.075 for \textit{Burgers'}  and the long range time is respectively 400 and 0.15. Recall that $\Delta t_{adv} = 1$ and  $\Delta t_{bur} = 0.0005$. All relative errors are in percentages.
    } 
    \centering
    \resizebox{\textwidth}{!}{  
    \begin{tabular}{lllllll}
\toprule
        &  &\bf Model & \bf \makecell[l]{Test relative Error} &\bf \makecell[l]{Relative error at mid-range \\ $T_{adv} = 40 \cdot \Delta t_{adv}$ \\ $T_{bur} = 150 \cdot \Delta t_{bur}$} &\bf \makecell[l]{Relative error at long-range \\ $T_{adv} = 400 \cdot \Delta t_{adv}$ \\ $T_{bur} = 300 \cdot \Delta t_{bur}$} \\
        \midrule
      \multirow{3}{*}{\rotatebox[origin=c]{45}{\bf \textit{Advection}}} &\multirow{3}{*}{\rotatebox[origin=c]{15}}
       & Explicit Res Net & \textbf{0.5 ± 0.009} & 106.5 ± 66.0 & $+ \infty$ \\
       & & FNO  & 1.1 ± 0.1 & \textbf{34.5 ± 19.0} & $7.2 \cdot 10^{5}$ ± $1.6 \cdot 10^{6}$ \\
       & & Implicit ResNet (Ours)  & 10.7 ± 4.2 & 1026.8 ± 346.7 & \textbf{1037.1 ± 347.9} \\
        \midrule
        \multirow{3}{*}{\rotatebox[origin=c]{45}{\bf \textit{Burgers'}}} &\multirow{3}{*}{\rotatebox[origin=c]{15}}
       & Explicit Res Net & 2.9 ± 0.3 & $6.9 \cdot 10^{10}$ ± $1.5 \cdot 10^{11}$  & $+ \infty$  \\
       & & FNO  & \textbf{0.6 ± 0.004} & $3.6 \cdot 10^{6}$ ± $8.0 \cdot 10^{6}$ & $+ \infty$ \\
       & & Implicit ResNet (Ours)  & 10.7 ± 0.3 & \textbf{277.7 ± 102.6} & \textbf{340.0 ± 129.9}\\
    \bottomrule
    \end{tabular}
    }
    \label{tab:rel_main_res}
\end{table}

\end{document}